%% file: main.tex
\documentclass[10pt,twocolumn,letterpaper]{article}
\usepackage{temp}
\usepackage{times}
\usepackage{epsfig}
\usepackage{graphicx}
\usepackage{amsmath}
\usepackage{amssymb}
\usepackage{algorithm}
\usepackage{algorithmic}
\usepackage{subcaption}
\usepackage{booktabs}
\usepackage{xspace}
\usepackage[pagebackref=true,breaklinks=true,letterpaper=true,colorlinks,bookmarks=false]{hyperref}

\begin{document}

\newcommand{\algorithmautorefname}{Alg.}

\title{Waving Goodbye to Low-Res: \\
A Diffusion-Wavelet Approach for Image Super-Resolution}

\author{Brian B. Moser$^{1,2}$, Stanislav Frolov$^{1,2}$, Federico Raue$^{1}$, Sebastian Palacio$^{1}$, Andreas Dengel$^{1,2}$\\
$^1$ German Research Center for Artificial Intelligence (DFKI), Germany\\
$^2$ RPTU Kaiserslautern-Landau, Germany\\
{\tt\small first.second@dfki.de}
}

\maketitle

\begin{abstract}
   This paper presents a novel Diffusion-Wavelet (DiWa) approach for Single-Image Super-Resolution (SISR). It leverages the strengths of Denoising Diffusion Probabilistic Models (DDPMs) and Discrete Wavelet Transformation (DWT). 
   By enabling DDPMs to operate in the DWT domain, our DDPM models effectively hallucinate high-frequency information for super-resolved images on the wavelet spectrum, resulting in high-quality and detailed reconstructions in image space. 
   Quantitatively, we outperform state-of-the-art diffusion-based SISR methods, namely SR3 and SRDiff, regarding PSNR, SSIM, and LPIPS on both face (8x scaling) and general (4x scaling) SR benchmarks.
   Meanwhile, using DWT enabled us to use fewer parameters than the compared models: 92M parameters instead of 550M compared to SR3 and 9.3M instead of 12M compared to SRDiff.
   Additionally, our method outperforms other state-of-the-art generative methods on classical general SR datasets while saving inference time. 
   Finally, our work highlights its potential for various applications.
\end{abstract}

\section{Introduction}
Single-Image Super-Resolution (SISR) is an ill-posed and long-standing challenge in computer vision since many High-Resolution (HR) images can be valid for any given Low-Resolution (LR) image.
While regression-based methods like Convolutional Neural Networks (CNNs) may work at low magnification ratios, they often fail to reproduce the high-frequency details needed for high magnification ratios.
Generative models and, more recently, Denoising Diffusion Probabilistic Models (DDPMs) have proven to be effective tools for such cases \cite{whang2022deblurring, dos2022face, chung2022mr}. 
Moreover, DDPMs produce reconstructions with subjectively perceived better quality compared to regression-based methods \cite{saharia2022image}. 
Further advancing DDPMs requires finer high-frequency details prediction \cite{10041995}. 
Another pressing demand is accessibility due to computationally-intensive requirements of DDPMs \cite{ganguli2022predictability}.

This work proposes a novel Diffusion-Wavelet approach (DiWa) that addresses both issues and leverages the capabilities of DDPMs by incorporating Discrete Wavelet Transformation (DWT).
We modify the DDPM pipeline to better learn high-frequency details by operating in the DWT domain instead of the image space. 
Our motivation for using DWT is two-fold: 
Firstly, combining DWT and DDPMs can improve image quality by enabling the model to capture and preserve essential features that may be lost or distorted when processed directly.
DWT provides an alternative representation, explicitly isolating high-frequency details in separate sub-bands. 
As a result, their representations are more sparse and, therefore, easier for a network to learn \cite{mallat1999wavelet, he2016deep}.
This property has also been exploited in diffusion-based audio synthesis with impressive results \cite{kong2020diffwave}.

Secondly, DWT halves the image's spatial size per the Nyquist rule \cite{guo2017deep}, which speeds up the inference time of the denoising function (CNN) and is particularly advantageous when the model is applied numerous times during DDPM inference.
In a recent work by Phung et al. \cite{phung2022wavelet}, a similar approach was employed for image generation using DiffusionGAN \cite{xiao2021tackling}, showcasing its speed-up potential. 
However, DiffusionGAN differs from traditional DDPMs by approximating intermediate steps with a GAN to reduce the required time steps for image generation.

Altogether, DWT offers tangible advantages to current methods in the field of SR.
Our work makes the following key contributions:
\begin{itemize}
 \item The first SR application of DDPMs in fusion with DWT for iterative high-frequency refinement that benefits from dimensional-reduced frequency representation in terms of performance and parameters.
 \item Our approach outperforms state-of-the-art diffusion models SR3 \cite{saharia2022image} for face-only SR and SRDiff \cite{li2022srdiff} as well as other generative approaches, namely SFTGAN \cite{wang2018recovering}, SRGAN \cite{ledig2017photo}, ESRGAN \cite{wang2018esrgan}, NatSR \cite{soh2019natural}, and SPSR \cite{ma2020structure}, on general SR.
 The improved performance makes our approach attractive as a pre-processing step for other applications like image classification \cite{zhang2018unreasonable}.
 \item The frequency-based representation, which has a 4x smaller spatial area compared to image space, enables a reduction of the denoise function's parameters since a smaller receptive field is required. It leads to a more accessible approach for researchers without access to large-scale computing resources: 
 Our approach needs 92M parameters instead of 550M compared to SR3, and 9.3M instead of 12M compared to SRDiff.
\end{itemize}

\section{Background}
Fusing Discrete Wavelet Transformation (DWT) and Denoising Diffusion Probabilistic Models (DDPMs) exploits the generative power of DDPMs while using the representation benefits of DWT.
This results in faster inference, sparser learning targets and, therefore, sharper results.
This section lays out the definitions of 2D-DWT and DDPMs.

\subsection{Discrete Wavelet Transformation}
Discrete Wavelet Transformation (DWT) is a technique for analyzing and representing signals that can reveal important details that are not apparent in raw data. 
DWT is used in various image processing applications, including image denoising, compression and SR \cite{thompson1981digital, akansu2001multiresolution, liu2018multi}.

Given a signal $\mathbf{x} \left[ n \right] \in \mathbb{R}^N$, the 1D Discrete Wavelet Transformation (1D-DWT) first applies a half band high-pass filter $h \left[ n \right]$ and then a low-pass filter $l\left[ n \right]$.
The formulation of the filters depends on the wavelet choice.
A well-known choice is the Haar (``db1''') wavelet \cite{haar1911theorie}.

Every image can be represented as a two-dimensional signal with index $\left[ n, m\right]$, where $n$ and $m$ represent the columns and rows, respectively.
Hence, let $\mathbf{x} \in \mathbb{R}^{w \times h \times c}$ be a color image. 2D-DWT captures the image details in four sub-bands: average ($A$), vertical ($V$), horizontal ($H$), and diagonal ($D$) information:
\begin{equation}
    \check{\mathbf{x}} = \text{2D-DWT} \left( \mathbf{x} \right) = \mathbf{x}_A \odot  \mathbf{x}_V \odot \mathbf{x}_H \odot \mathbf{x}_D,
    \label{eq:dwt}
\end{equation}
\noindent
where $\odot$ is a channel-wise concatenation operator and $\mathbf{x}_A$ are the average, $\mathbf{x}_V$ the vertical, $\mathbf{x}_H$ the horizontal and $\mathbf{x}_D$ the diagonal details of $\mathbf{x}$, respectively, with $\mathbf{x}_A, \mathbf{x}_V, \mathbf{x}_H, \mathbf{x}_D \in \mathbb{R}^{\frac{w}{2} \times \frac{h}{2} \times c}$.

The 2D-IDWT is the inverse operation of 2D-DWT:
\begin{equation}
    \text{2D-IDWT} \left( \check{\mathbf{x}} \right) = \mathbf{x}.
\end{equation}
\subsection{Denoising Diffusion Probabilistic Models}
The core idea behind Denoising Diffusion Probabilistic Models (DDPMs) is to iteratively update the pixel values based on their neighbors \cite{ho2020denoising}. 
This process can be thought of as ``diffusing'' the information in the image over time to smooth out noise and reduce the overall variability in the data. 
The application of DDPMs is generally divided into two phases: the forward diffusion and the reverse process. 

The diffusion process gradually adds noise to the input, while the reverse process aims to denoise the added noise iteratively. 
Through iterative refinement, it is easier for a CNN to keep track of small perturbations and correct them instead of predicting one large transformation \cite{ho2020denoising, nichol2021improved}.

Let $\mathcal{D} = \{(\mathbf{x}_i, \mathbf{y}_i)\}_{i=1}^N$ be a dataset with unknown conditional distribution $p(\mathbf{y} \,|\, \mathbf{x})$.
For SR, $\mathbf{x}_i$ is the LR image while $\mathbf{y}_i$ is the corresponding HR image; between $\mathbf{x}_i$ and $\mathbf{y}_i$ is an unknown degradation relationship.

\subsubsection{Gaussian Diffusion Process}
To enable an iterative refinement, the Markovian diffusion process $q$ adds Gaussian noise to an input image  $\mathbf{z}_{0} \leftarrow \mathbf{x}$ over $T$ steps \cite{nichol2021improved, sohl2015deep}:
\begin{align}
    q(\mathbf{z}_{1:T} \mid \mathbf{z}_{0}) &= \prod\nolimits_{t=1}^{T} q(\mathbf{z}_{t} \mid \mathbf{z}_{t-1}), \\
    q(\mathbf{z}_{t} \mid \mathbf{z}_{t-1}) &= \mathcal{N}(\mathbf{z}_{t} \mid \sqrt{\alpha_t}\, \mathbf{z}_{t-1}, (1 - \alpha_t) \mathbf{I} )
\end{align}
\noindent
with $0 < \alpha_{1:T} < 1$ as added noise variance per time step hyper-parameter.
A major benefit of this formulation is that it can be reduced to a single-step calculation by
\begin{equation}
    q(\mathbf{z}_t \mid \mathbf{z}_0) = \mathcal{N}(\mathbf{z}_t \mid \sqrt{\gamma_t}\, \mathbf{z}_0, (1-\gamma_t) \mathbf{I}),
\end{equation}
where $\gamma_t = \prod_{i=1}^t \alpha_i$. 
This marginalization enables a one time step training for an arbitrary $t \in \{1, ..., T \}$.
\subsubsection{Conditional Reverse Process}
The reverse Markovian process $p$ starts from Gaussian noise $\mathbf{z}_{T}$ and performs the inference conditioned on $\mathbf{x}$ by
\begin{align}
    p_\theta(\mathbf{z}_{0:T} | \mathbf{x}) &= p(\mathbf{z}_T) \prod\nolimits_{t=1}^T p_\theta(\mathbf{z}_{t-1} | \mathbf{z}_t, \mathbf{x}) \\
    p(\mathbf{z}_T) &= \mathcal{N}(\mathbf{z}_T \mid \mathbf{0}, \mathbf{I})\\
    p_\theta(\mathbf{z}_{t-1} | \mathbf{z}_{t}, \mathbf{x}) &= \mathcal{N}(\mathbf{z}_{t-1} \mid \mu_{\theta}(\mathbf{x}, \mathbf{z}_{t}, \gamma_t), \sigma_t^2\mathbf{I}).
\end{align}
The mean $\mu_{\theta}$ depends on a parameterized denoising function $f_\theta$, which can either predict the added noise $\varepsilon$ or the underlying image $\mathbf{z}_0$. Following the standard approach of Ho et al. \cite{ho2020denoising}, we focus on predicting the noise in this work. 
Therefore, the mean is
\begin{equation}
    \mu_{\theta}(\mathbf{x}, \mathbf{z}_{t}, \gamma_t) = \frac{1}{\sqrt{\alpha_t}} \left( \mathbf{z}_{t} - \frac{1-\alpha_t}{\sqrt{1 - \gamma_t}} f_\theta \left( \mathbf{x}, \mathbf{z}_{t}, \gamma_t \right)\right).
\end{equation}
Following Saharia et al. \cite{saharia2022image}, setting the variance of $p_\theta(\mathbf{z}_{t-1}|\mathbf{z}_t, \mathbf{x})$ to $(1 - \alpha_t)$ results in the following refining step:
\begin{equation}
    \mathbf{z}_{t-1} \leftarrow \frac{1}{\sqrt{\alpha_t}} \left( \mathbf{z}_{t} - \frac{1-\alpha_t}{\sqrt{1 - \gamma_t}} f_\theta \left( \mathbf{x}, \mathbf{z}_{t}, \gamma_t \right)\right) + \sqrt{1 - \alpha_t} \varepsilon_t,
\end{equation}
\noindent
where $\varepsilon_t \sim \mathcal{N}(\textbf{0},\,\textbf{I})$.

\begin{figure}[t]
    \begin{center}
        \includegraphics[width=0.40\textwidth]{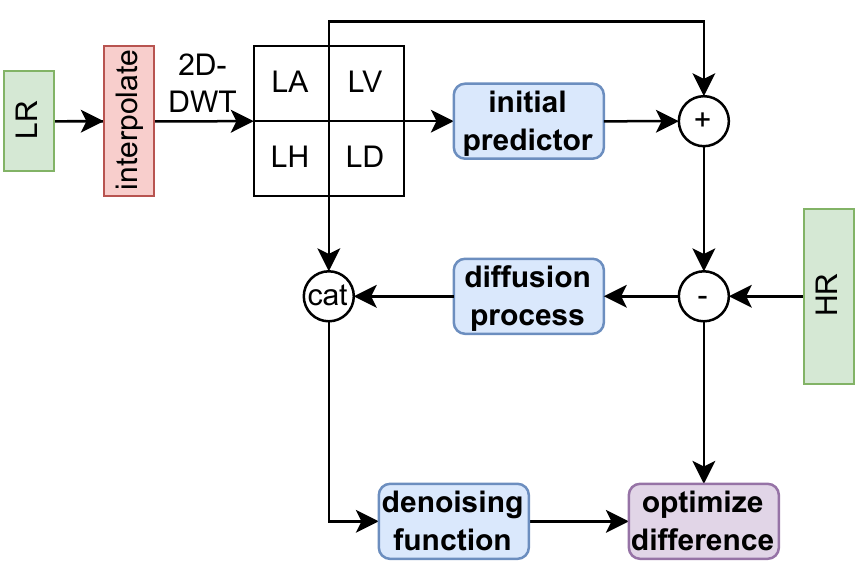}
        \caption{ \label{fig:training}
        Overview of training.  
        The diffusion process takes the difference between the initial predictor and the corresponding HR image as input.
        The trained reverse process learns to denoise the noisy residual image with the difference between the reconstruction of the initial predictor and the corresponding HR image as the optimization target.
        }
    \end{center}
\end{figure}

\input{algorithms/training}

\section{Methodology}
Our proposed Diffusion-Wavelet approach (DiWa) combines conditional diffusion models with wavelet decomposition to enable sparser and easier learning targets for faster inference and finer reconstructions. 
Hence, our diffusion model operates on the wavelet spectrum instead of the original image space.
It takes advantage of the unique properties of the wavelet domain to exploit high-frequency information and improve the quality of the final reconstruction.

We begin this section by introducing how we used the 2D-DWT domain. 
Then we describe how we integrate and optimize an initial predictor, which generates an initial estimate of the final reconstruction.
We provide the overall algorithm of our method in \autoref{alg:training} (training) and \autoref{alg:inference} (inference).
It consists of three main components: the 2D-DWT representation, the initial predictor, and the  diffusion.

\begin{figure}[t]
    \begin{center}
        \includegraphics[width=0.40\textwidth]{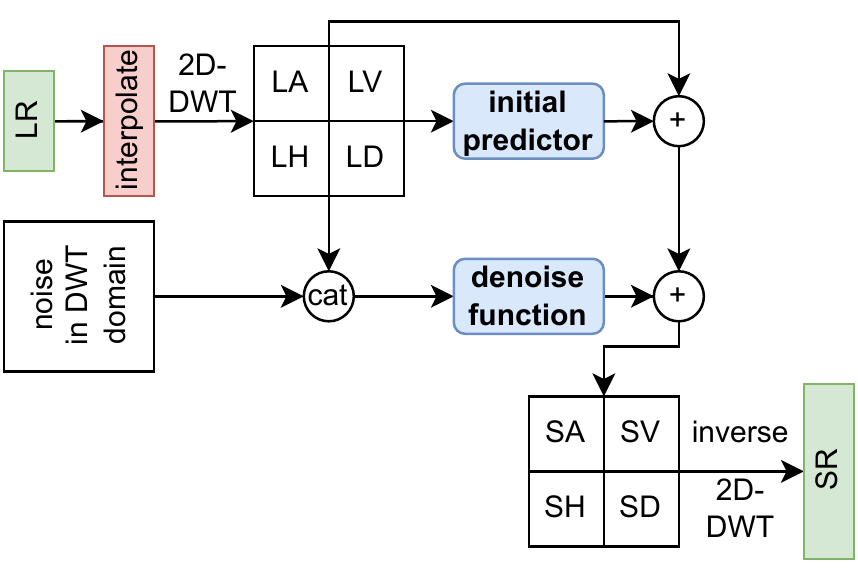}
        \caption{ \label{fig:inference}
        Overview of inference. 
        The image is first decomposed into sub-bands using 2D-DWT, which an initial predictor processes.
        The denoise function then adds and computes the remaining details by conditioning on the sub-bands, incorporating noise.
        The result is then returned to the pixel domain via the inverse 2D-DWT function.
        }
    \end{center}
\end{figure}

\input{algorithms/inference}

\subsection{DWT Domain}
Let $(\mathbf{x}_i, \mathbf{y}_i) \in \mathcal{D}$ be an LR-HR image pair. 
Before applying the diffusion process, we translate $\mathbf{x}_i$ to the wavelet domain via $\check{\mathbf{x}}_i = \text{2D-DWT} \left( \mathbf{x}_i \right)$ and Haar ("db1") \cite{haar1911theorie} wavelets.
The decomposition separates the input into four sub-bands, representing the average image (LA) and high-frequency details in the horizontal, vertical, and diagonal direction (LH, LV, LD) as described in \autoref{eq:dwt}.
Meanwhile, the spatial area of the sub-bands is four times smaller than the original image.
Next, Gaussian noise $\varepsilon \sim \mathcal{N}(\textbf{0},\,\textbf{I})$ is added to the sub-bands, and the denoise function $f_\theta$ learns to operate in that domain.
We also optimize the parameters $\theta$ of the denoise function in the wavelet domain with target $\check{\mathbf{y}}_i$.
For final inference sampling, our method returns the inversed transformation 2D-IDWT.

By operating in the wavelet domain, we open the possibility for the denoise function $f_\theta$ to focus on isolated high-frequency details, often lost when the data is processed directly.
Also, utilizing DWT allows for faster inference times as the spatial size is halved due to the Nyquist rule \cite{guo2017deep}.

\subsection{Initial Predictor}
Due to different variances in the DWT sub-bands, the denoise function must adapt to different distributions.
In particular, the average sub-band is similar to a downsampled version of the original image.
Therefore, it holds richer information than the sparse representations of the remaining high-frequency sub-bands.
We overcome the adaptation by learning only the residuals, resulting in more comparable target distributions.
Thus, our method refines the output of a deterministic initial predictor $g_\theta$ that provides a diverse yet plausible SR estimation of a given LR input.
This further pushes the paradigm of high-frequency prediction by learning the sparse and missing details that traditional SR models fail to capture.
The initial predictor is inspired by Whang et al. \cite{whang2022deblurring} but operates in the wavelet domain.

Since our method works in the 2D-DWT domain and for simplicity, we use the DWSR network from Guo et al. \cite{guo2017deep} as the initial predictor $g_\theta$, a simple 10-layer wavelet CNN for SR that predicts the four HR wavelet coefficients.

\subsection{Optimization}

The denoise function $f_\theta$ is optimized to remove the added noise. Thus, we minimize the objective function
\begin{equation}
    \mathcal{L} \left( \theta \right) = \mathop{\mathbb{E}}_{(\mathbf{x}, \mathbf{y})} \mathop{\mathbb{E}}_{t} \bigg\lVert \varepsilon_t - f_\theta \left( \check{\mathbf{x}}, \mathbf{z}_{t}, \gamma_t \right) \bigg\rVert_1
\label{eq:loss}
\end{equation}
\begin{equation}
    \mathbf{z}_{t} = \sqrt{\gamma_t} \left( \check{\mathbf{y}} - g_\theta \left( \check{\mathbf{x}} \right)\right) + \sqrt{1-\gamma_t}\varepsilon_t
\label{eq:lossell}
\end{equation}
and the initial predictor is part of the objective function like in Whang et al. \cite{whang2022deblurring}. 
Consequently, an additional or auxiliary loss function to train the initial predictor is unnecessary since the gradients $\nabla_{\theta} f_\theta$ flow through $f_\theta$ into $\nabla_{\theta} g_\theta$.

The parameterized denoise function $f_\theta$ only needs to model the residual sub-bands, which poses an easier learning target than reconstructing the entire image.
Therefore, the initial predictor improves the overall efficiency of the optimization process and reduces the number of iterations needed to converge to a satisfactory solution.
Inference and training are depicted in \autoref{fig:training} and \autoref{fig:inference}.


\section{Experiments}
We evaluate our proposed method for general SR as well as for the more challenging scenario of face SR.
Our experiments aim to demonstrate our approach's effectiveness and compare its performance to SR3 \cite{saharia2022image} for face SR and SRDiff \cite{li2022srdiff} alongside other state-of-the-art generative approaches \cite{lugmayr2020srflow, ma2020structure, lim2017enhanced, soh2019natural, ledig2017photo, zhang2019ranksrgan, wang2018recovering, wang2018esrgan} for general SR.
We present visual examples, as well as quantitative and qualitative results for both tasks.
Overall, we achieved high-quality results for the face and general SR task and outperformed the compared methods on standard metrics PSNR, SSIM, and LPIPS.

\subsection{Datasets}
For face SR, we evaluate SRDiff against our method. 
For general SR, we benchmark our approach against SR3.
As shared in the literature, we employed bicubic interpolation and anti-aliasing procedures to generate LR-HR image pairs, which discards high-frequency information \cite{MATLAB:2017b}.

\textbf{Face SR:} We used Flickr-Faces-HQ (FFHQ) \cite{karras2019style}, 50K high-quality face images from Flickr, as training. 
For evaluation, we utilized CelebA-HQ \cite{karras2017progressive}, which consists of 30K face images.
We followed two 8x scaling tracks like in Saharia et al. \cite{saharia2022image}.
We resized all images to match the cases $16\times16 \rightarrow 128\times128$ and $64\times64 \rightarrow 512\times512$.

\textbf{General SR:} We used 800 2K resolution high-quality images from DIV2K \cite{agustsson2017ntire} for training and the datasets Set5 \cite{bevilacqua2012low}, Set14 \cite{zeyde2010single}, BSDS100 \cite{martin2001database}, and General100 \cite{dong2016accelerating} for evaluation. 
In addition, we used the DIV2K validation set to compare our approach with SRDiff.
As common in the literature, we followed the standard procedure and extracted sub-images $48\times48 \rightarrow 192\times192$ from DIV2K for 4x scaling training \cite{dong2015image, 10041995}.
For testing, we kept the original sizes of the images, a standard procedure in SISR.

\subsection{Training Details}
The code and trained models for our experiments can be found on GitHub\footnote{\url{https://github.com/Brian-Moser/diwa}}, which complements the unofficial implementation of SR3\footnote{\url{https://github.com/Janspiry/Image-Super-Resolution-via-Iterative-Refinement}}.

\textbf{Diffusion-specific:} To achieve a fine-grained diffusion process during training, we set the time step to 2,000. 
We reduced the time steps to 500 during evaluation for faster inference, like in Saharia et al. \cite{saharia2022image}. 
We avoided other time step evaluations as they would affect comparability.
We trained for 1M iterations for face SR and 100k for general SR to train our models. 
The linear noise schedule has the endpoints of $1-\alpha_0 = 10^{-6}$ and $1-\alpha_T = 10^{-2}$. 

\textbf{Regularization: } All our experiments apply horizontal flipping with a probability of 50\%. 
In addition, we use dropout \cite{srivastava2014dropout} (with 10\%) in all experiments except for $64\times64 \rightarrow 512\times512$ face SR.

\textbf{Architecture:} Similar to SR3 and SRDiff, we employ a U-Net architecture \cite{ronneberger2015u} as a denoise function. 
In contrast to the approach adopted by SR3, we employed residual blocks \cite{he2016deep} proposed by Ho et al. \cite{ho2020denoising} instead of those used in BigGAN \cite{brock2018large}.
Contrary to SRDiff, we do not use a pre-trained LR encoder beforehand to convert LR images into feature representations but use an initial predictor instead.

\textbf{Optimizer:} For training, we utilized AdamW \cite{loshchilov2017decoupled} optimizer with a weight decay of $10^{-4}$ instead of Adam \cite{kingma2014adam}.
Unlike other diffusion-based approaches, we do not employ exponential moving average (EMA) over model parameters \cite{nichol2021improved} to save additional computation.
Our method outperformed SR3 and SRDiff without EMA, even though it is very effective in improving the quality of DDPMs. 

\textbf{Face SR details:}
\label{sec:fsr-details}
In our $16\times16 \rightarrow 128\times128$ face SR experiments, we employed a smaller setup than SR3. We reduced the channel dimension to 64 instead of 128 and the number of ResNet Blocks to 2 in place of 3, resulting in a total of roughly 92M parameters instead of 550M. For $64\times64 \rightarrow 512\times512$ face SR, we adopted the same architecture settings as SR3 (625M parameters) to provide a fair subjective comparison for visual examples. The learning rate was set to $1\times10^{-4}$. Additionally, the batch size was also reduced to 4 rather than 256, which was necessary to run the experiments on A100 GPUs.

\textbf{General SR Details:}
Our comparison with SRDiff employed a smaller yet comparable architecture, with channel multipliers of [1, 2, 2, 4], a channel size of 48, and two ResNet blocks, resulting in a model with 9.3M parameters, as opposed to SRDiff's 12M. In addition, the reduced model size enabled us to train with a larger batch size of 256. 
The learning rate was set to $2\times10^{-5}$.

\subsection{Results}
This section presents our proposed method's quantitative and qualitative results for face and general SR. 
We compare our performance to state-of-the-art diffusion methods, SR3 and SRDiff, using standard metrics such as Peak Signal-to-Noise Ratio (PSNR), Structural Similarity Index (SSIM), and Learned Perceptual Image Patch Similarity (LPIPS)  with AlexNet \cite{10041995, zhang2018unreasonable, krizhevsky2017imagenet}. 
Higher PSNR and SSIM values indicate better image quality. In contrast, a lower LPIPS value indicates better-perceived quality.

\begin{table}[!t]
\begin{center}
    \begin{tabular}{l  c c c  c}
    \toprule
     & Pulse& FSRGAN& SR3& DiWa\\
     & \cite{menon2020pulse} & \cite{chen2018fsrnet} & \cite{saharia2022image} & (ours)\\
    \midrule
    \textbf{PSNR} $\uparrow$ & 16.88 & 23.01 & 23.04 & \textbf{23.34}\\
    \textbf{SSIM} $\uparrow$ & 0.44  & 0.62 & 0.65 & \textbf{0.67}\\
    \bottomrule
    \end{tabular}
    \caption{\label{tab:psnr_ssim_faces} PSNR and SSIM comparison on 16$\times$16 $\rightarrow$ 128$\times$128 face SR (CelebA-HQ). Our method outperforms SR3 in both metrics while having roughly 458M fewer parameters and fewer images during train iterations. Numbers are provided by Saharia et al. \cite{saharia2022image}.}
\end{center}
\end{table}

\begin{figure}[!t] 
   \begin{subfigure}{0.156\textwidth}
       \includegraphics[width=\linewidth]{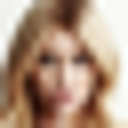}
       \caption{LR}
       \label{fig:subim1}
   \end{subfigure}
\hfill 
   \begin{subfigure}{0.156\textwidth}
       \includegraphics[width=\linewidth]{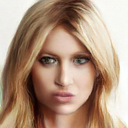}
       \caption{SR}
       \label{fig:subim2}
   \end{subfigure}
\hfill 
   \begin{subfigure}{0.156\textwidth}
       \includegraphics[width=\linewidth]{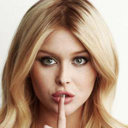}
       \caption{HR}
       \label{fig:subim3}
   \end{subfigure}

   \caption{
   A Comparison of a LR, SR, and HR image (CelebA-HQ) illustrates the quality of our proposed method for the $16\times16 \rightarrow 128\times128$ setting. The LR image shows a significant loss of information, particularly the presence of a finger in front of the mouth. Our proposed method can reconstruct the image with great detail, particularly in the hair. However, the HR image shows that our method cannot reconstruct the finger. Also, the HR image shows more defined edges and sharper details of the eyes. }
   \label{fig:image16by16}
\end{figure}

\begin{figure} 
   \begin{subfigure}{0.156\textwidth}
       \includegraphics[width=\linewidth]{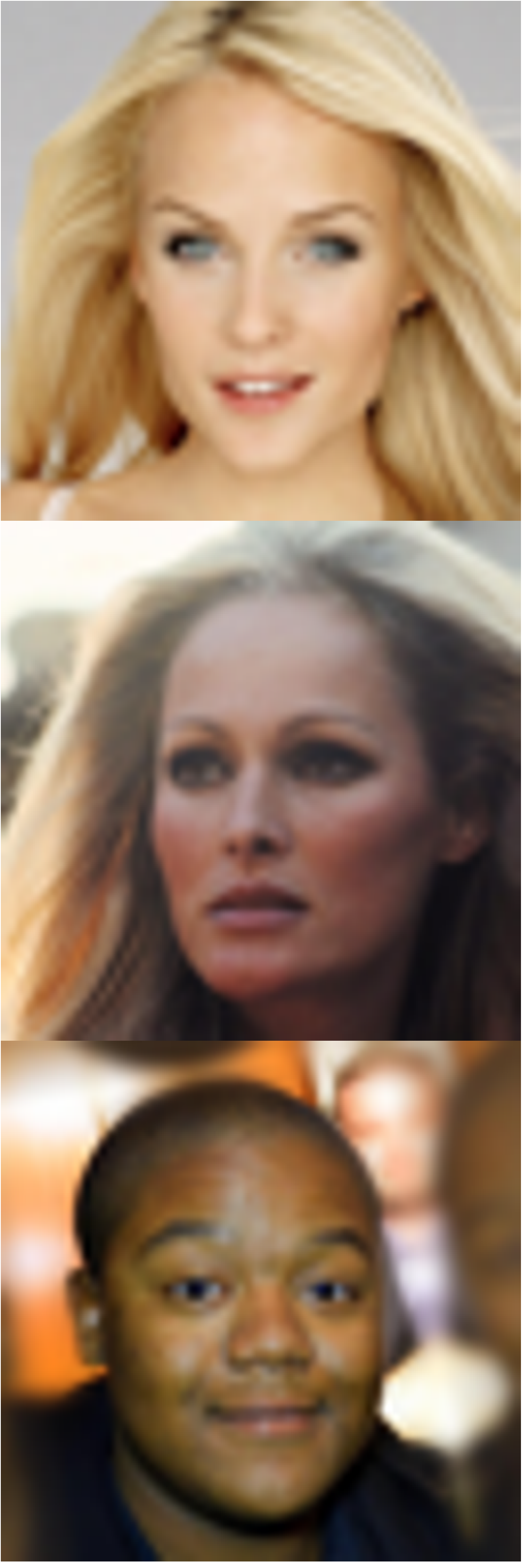}
       \caption{LR}
       \label{fig:subim1}
   \end{subfigure}
\hfill 
   \begin{subfigure}{0.156\textwidth}
       \includegraphics[width=\linewidth]{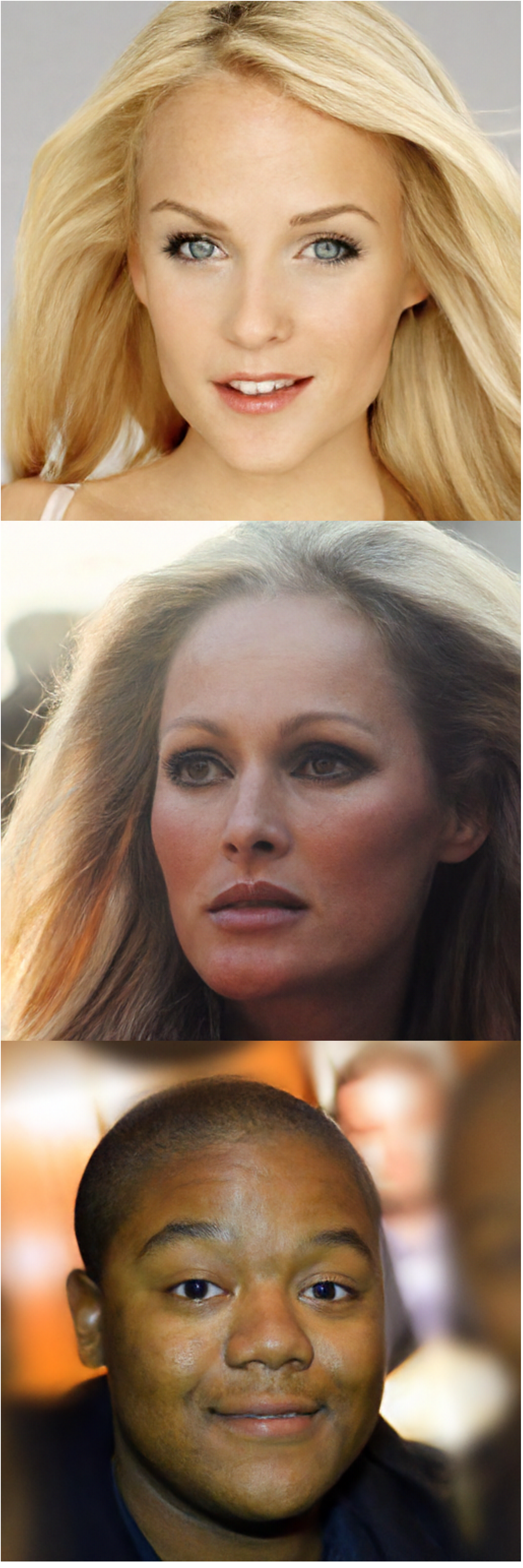}
       \caption{SR}
       \label{fig:subim2}
   \end{subfigure}
\hfill 
   \begin{subfigure}{0.156\textwidth}
       \includegraphics[width=\linewidth]{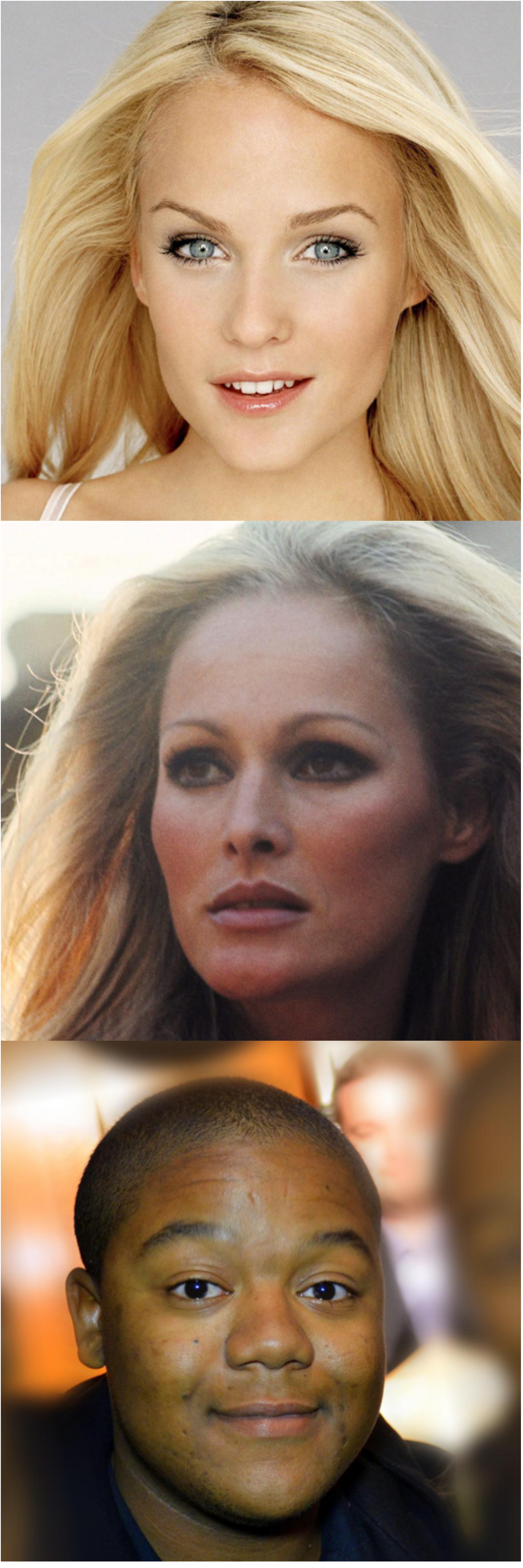}
       \caption{HR}
       \label{fig:subim3}
   \end{subfigure}

   \caption{LR, HR, and SR (our method) example results of the $64\times64 \rightarrow 512\times512$ experiments for three different face images (CelebA-HQ). The last row shows that our model produces continuous skin texture, which does not match the small details of the ground truth, such as moles and pimples.}
   \label{fig:image64by64}
\end{figure}

\subsubsection{Face Super-Resolution}
\autoref{tab:psnr_ssim_faces} shows that our approach outperforms SR3 and other generative methods applied to CelebA-HQ and in the 16$\times$16 $\rightarrow$ 128$\times$128 setting. 
Even though our model iterated over fewer samples (due to batch size) and has a significantly smaller parameter size (92M instead of 550M), it outperforms SR3 by 0.3 dB in terms of PSNR and 0.02 with regard to SSIM.
\autoref{fig:image16by16} visualizes example reconstruction images of our method. 
It demonstrates that using DWT is useful when the LR input image does not contain enough information (e.g., finger in the LR image) and additional details must be inferred or synthesized.

\begin{figure}
    \begin{center}
        \includegraphics[width=0.45\textwidth]{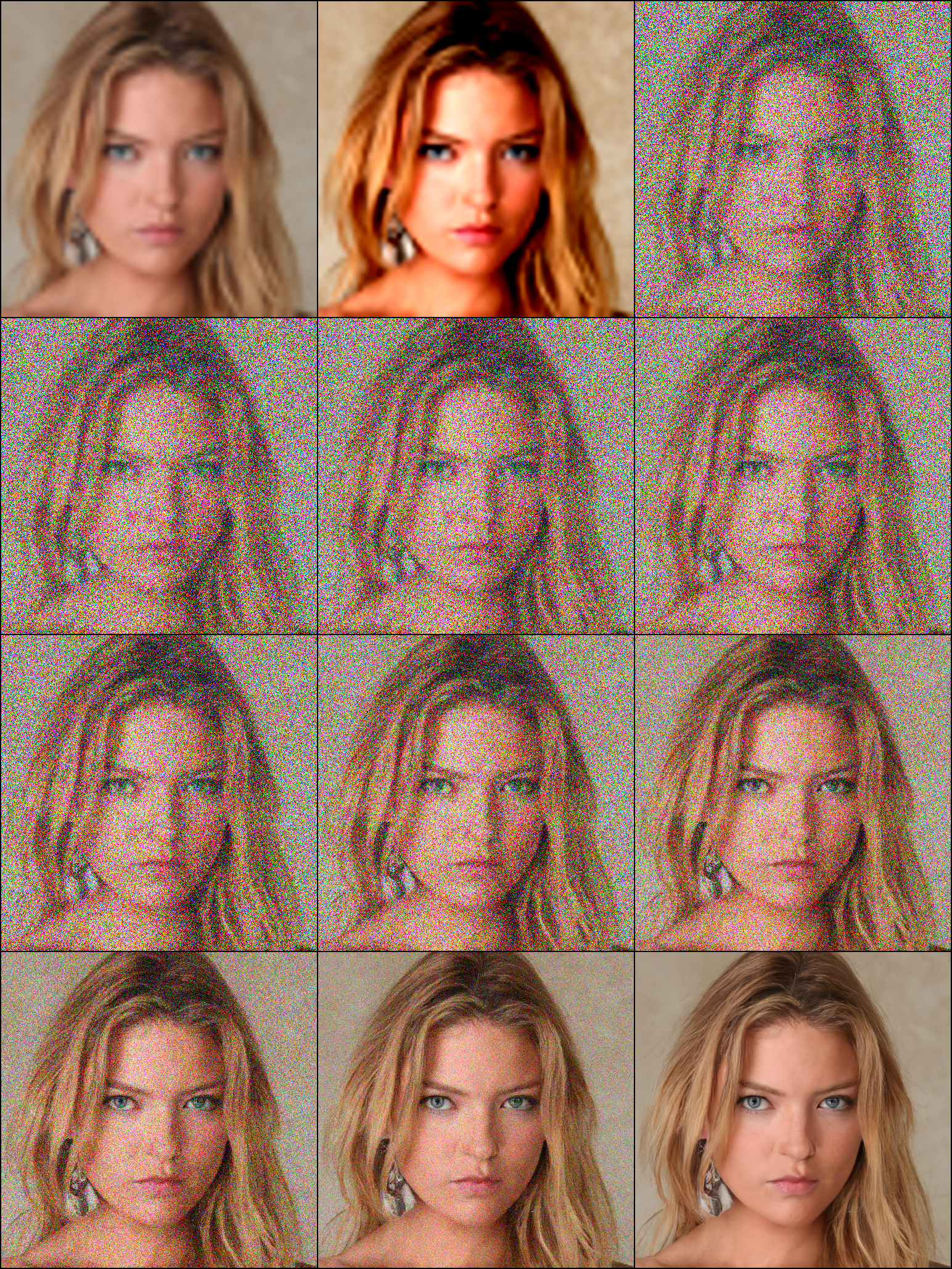}
        \caption{\label{fig:intermediate64}
        Intermediate denoising results were obtained with our approach on face super-resolution ($64\times64 \rightarrow 512\times512$). The top left image represents the LR input. The middle image in the first row is the estimation of our initial predictor. The remaining images show the intermediate denoising estimations from our denoising function as we apply it iteratively, progressing from left to right and top to bottom. The final prediction of the denoising function is in the lower right corner of the grid.}
    \end{center}
\end{figure}

\autoref{fig:image64by64} and \autoref{fig:intermediate64} illustrate the performance of our proposed method in the scenario of $64\times64 \rightarrow 512\times512$.
\autoref{fig:image64by64} showcases three example reconstructions and highlights the strengths and limitations of our method. 
Compared to the 16$\times$16 $\rightarrow$ 128$\times$128 reconstructions, it can produce more realistic high-quality HR results and exploits the additional information of the LR image more efficiently.
As observed, it produces smooth skin textures, similar to SR3, but struggles to preserve small details such as moles, pimples, or piercings. 

\autoref{fig:intermediate64} displays the intermediate denoising results and the reconstruction of the initial predictor for the $64\times64 \rightarrow 512\times512$ scenario. 
The results indicate that the initial predictor provides a strong baseline prediction but needs more nuanced details for high-resolution images. 
Therefore, it shows that high-frequency details are generated during diffusion.
Additionally, it can be noted that the colors in the initial prediction are saturated and contrast-rich in comparison to the ground truth.
Unfortunately, a direct comparison with SR3 is not possible for $64\times64 \rightarrow 512\times512$ as the authors do not provide quantitative results (PSNR or SSIM) in their paper. 
Due to the high hardware requirements for training an SR3 model with the exact specifications, we are unable to reproduce their results.

\begin{figure} 
   \begin{subfigure}{0.153\textwidth}
       \includegraphics[width=\linewidth]{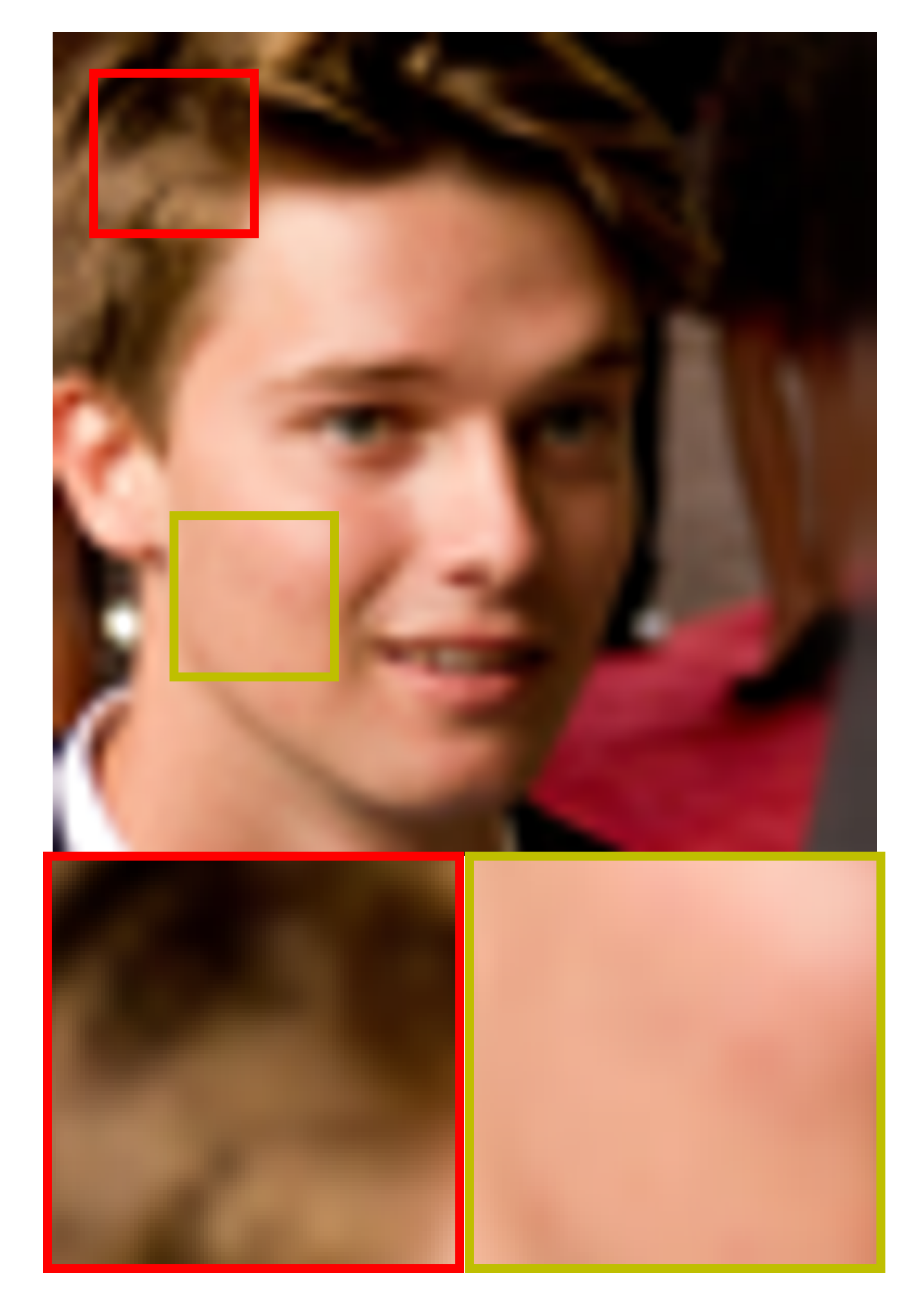}
       \caption{LR}
       \label{fig:subim1}
   \end{subfigure}
\hfill 
   \begin{subfigure}{0.153\textwidth}
       \includegraphics[width=\linewidth]{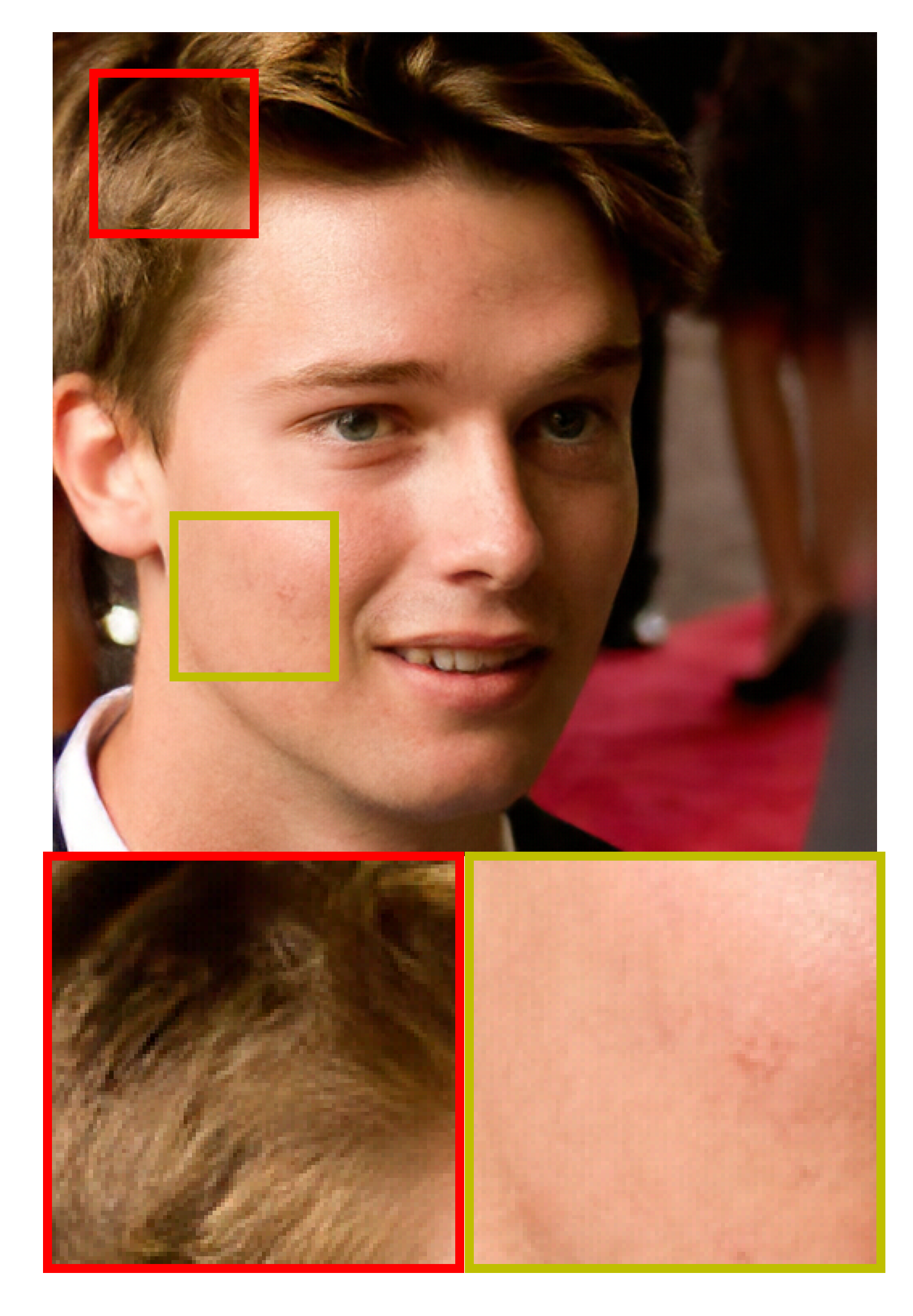}
       \caption{SR}
       \label{fig:subim2}
   \end{subfigure}
\hfill 
   \begin{subfigure}{0.153\textwidth}
       \includegraphics[width=\linewidth]{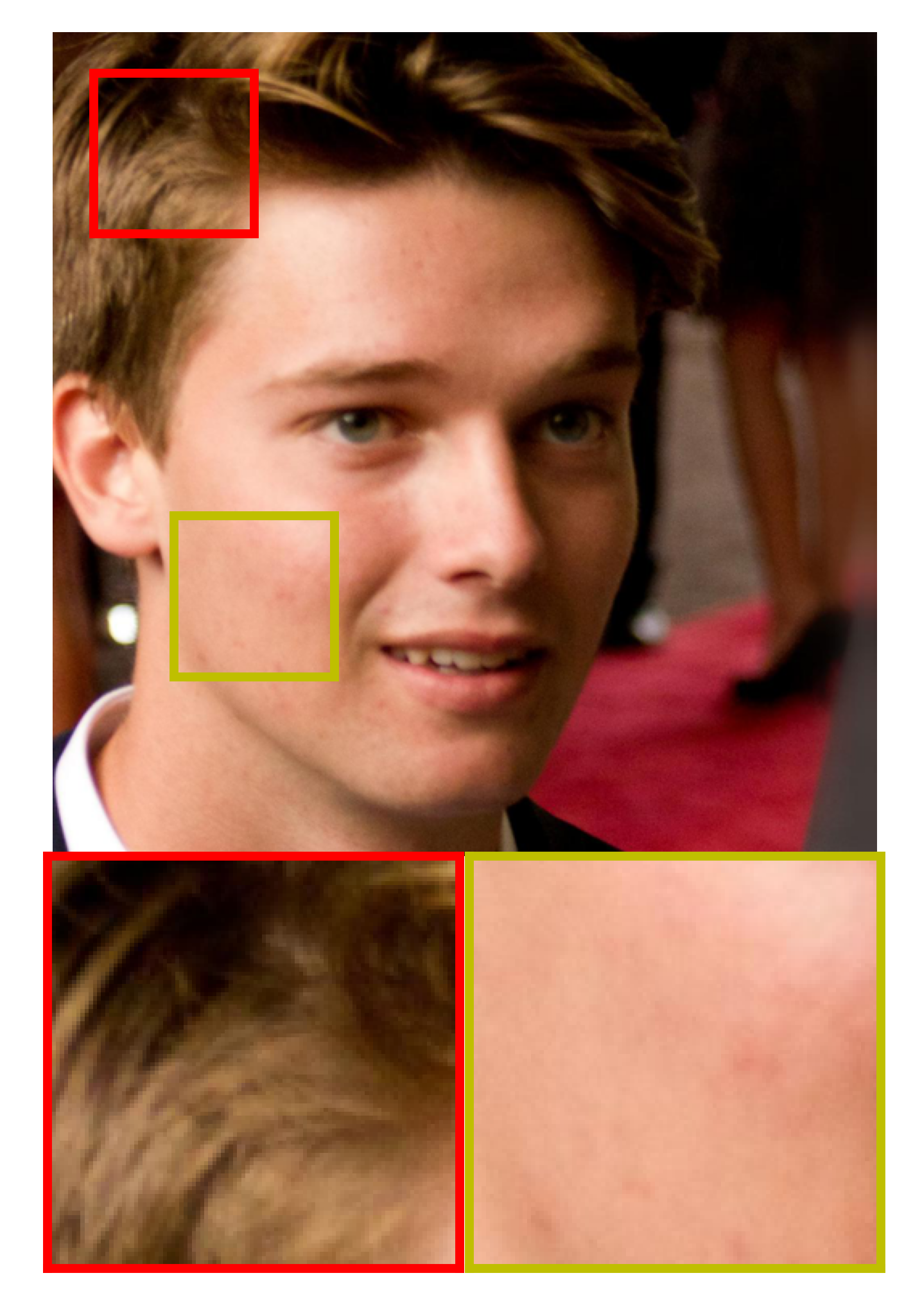}
       \caption{HR}
       \label{fig:subim3}
   \end{subfigure}

   \caption{A side-by-side comparison of zoomed-in regions of a LR, SR, and HR face image ($64\times64 \rightarrow 512\times512$, CelebA-HQ). The SR image, generated by our method, shows improved hair's fine details, with more strands and texture visible. However, compared to the HR image, it can be seen that the SR image falls short in terms of skin structure, specifically in depicting pimples and skin texture. }
   \label{fig:image64by64detail}
\end{figure}

The comparison of one face image with two zoomed-in regions is presented in \autoref{fig:image64by64detail}.
When examining the hair, one can see that our SR image has more delicate details than the HR image. 
The individual hair strands are more distinct and pronounced in the SR image. 
Similar to \autoref{fig:image64by64}, the HR image better represents the pores and pimples when looking at the skin structure.

\begin{table}
\center
\small
\begin{tabular}{l c c c }
\toprule
\textbf{Methods} & \textbf{PSNR} $\uparrow$ & \textbf{SSIM} $\uparrow$& \textbf{LPIPS} $\downarrow$ \\ 
\midrule
Bicubic & 26.70 & 0.77 & 0.409  \\
EDSR \cite{lim2017enhanced} & 28.98 & 0.83 & 0.270  \\
RRDB \cite{wang2018esrgan} & 29.44 & 0.84 & 0.253  \\ 
\midrule
RankSRGAN \cite{zhang2019ranksrgan} & 26.55 & 0.75 & 0.128  \\ 
ESRGAN \cite{wang2018esrgan} & 26.22 & 0.75 & 0.124  \\
SRFlow \cite{lugmayr2020srflow} & 27.09 & 0.76 & \underline{0.120}  \\
SRDiff \cite{li2022srdiff} & \underline{27.41} & \textbf{0.79} & 0.136  \\
DiWa (ours)  & \textbf{28.09} &  \underline{0.78} & \textbf{0.104}\\
\bottomrule
\end{tabular}
\caption{Results for 4× SR of general images on validation set of DIV2K. 
Our method outperforms all generative approaches concerning PSNR and LPIPS and achieves the second-best result \wrt SSIM. 
Note that EDSR and RRDB are regression-based methods that generally produce better PSNR and SSIM scores than generative approaches \cite{saharia2022image}. }
\label{tab:div2k_results}
\end{table}

\begin{figure*} 
   \begin{subfigure}{0.33\textwidth}
       \includegraphics[width=\linewidth]{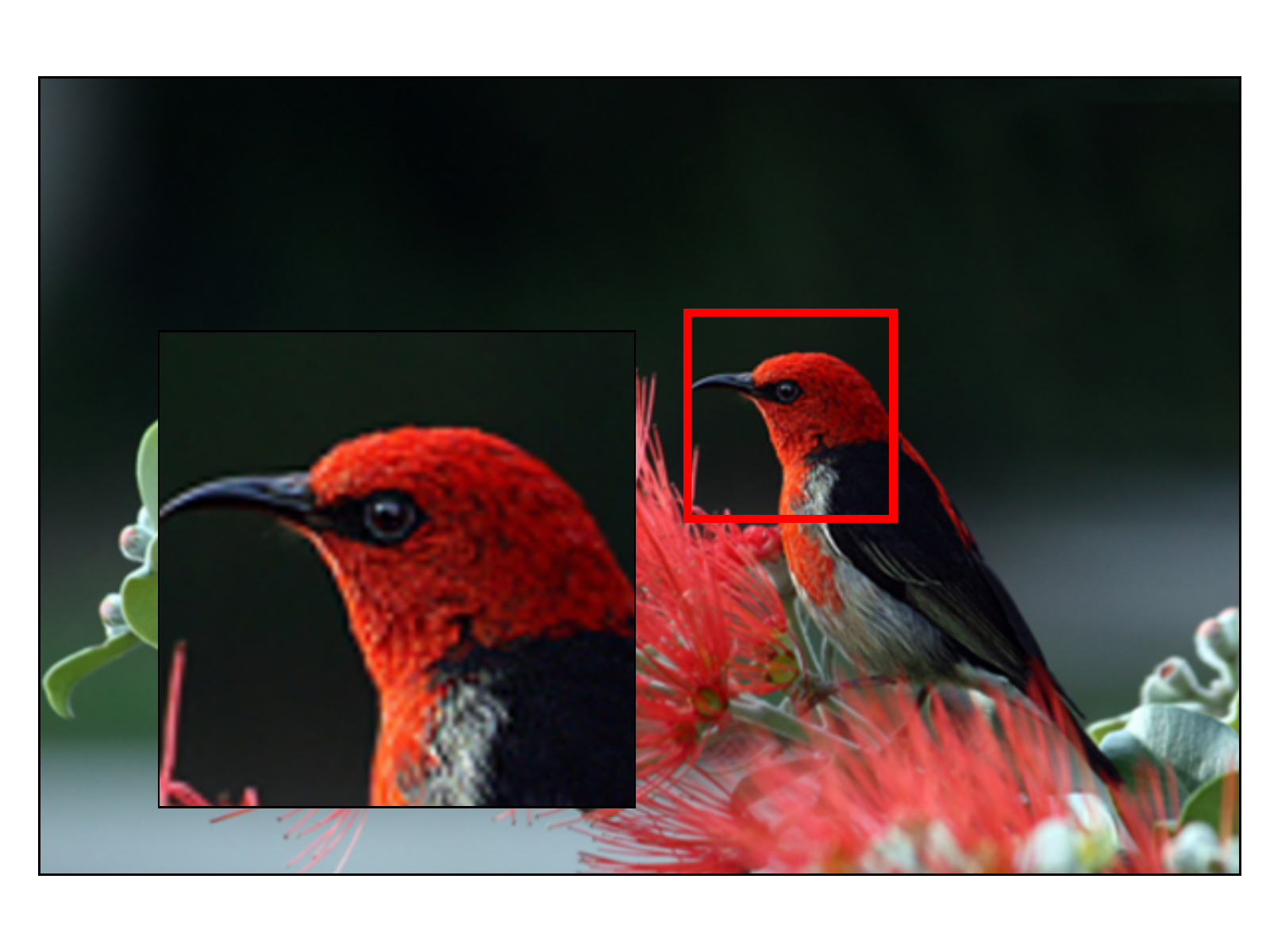}
       \caption{LR}
       \label{fig:subim1}
   \end{subfigure}
\hfill 
   \begin{subfigure}{0.33\textwidth}
       \includegraphics[width=\linewidth]{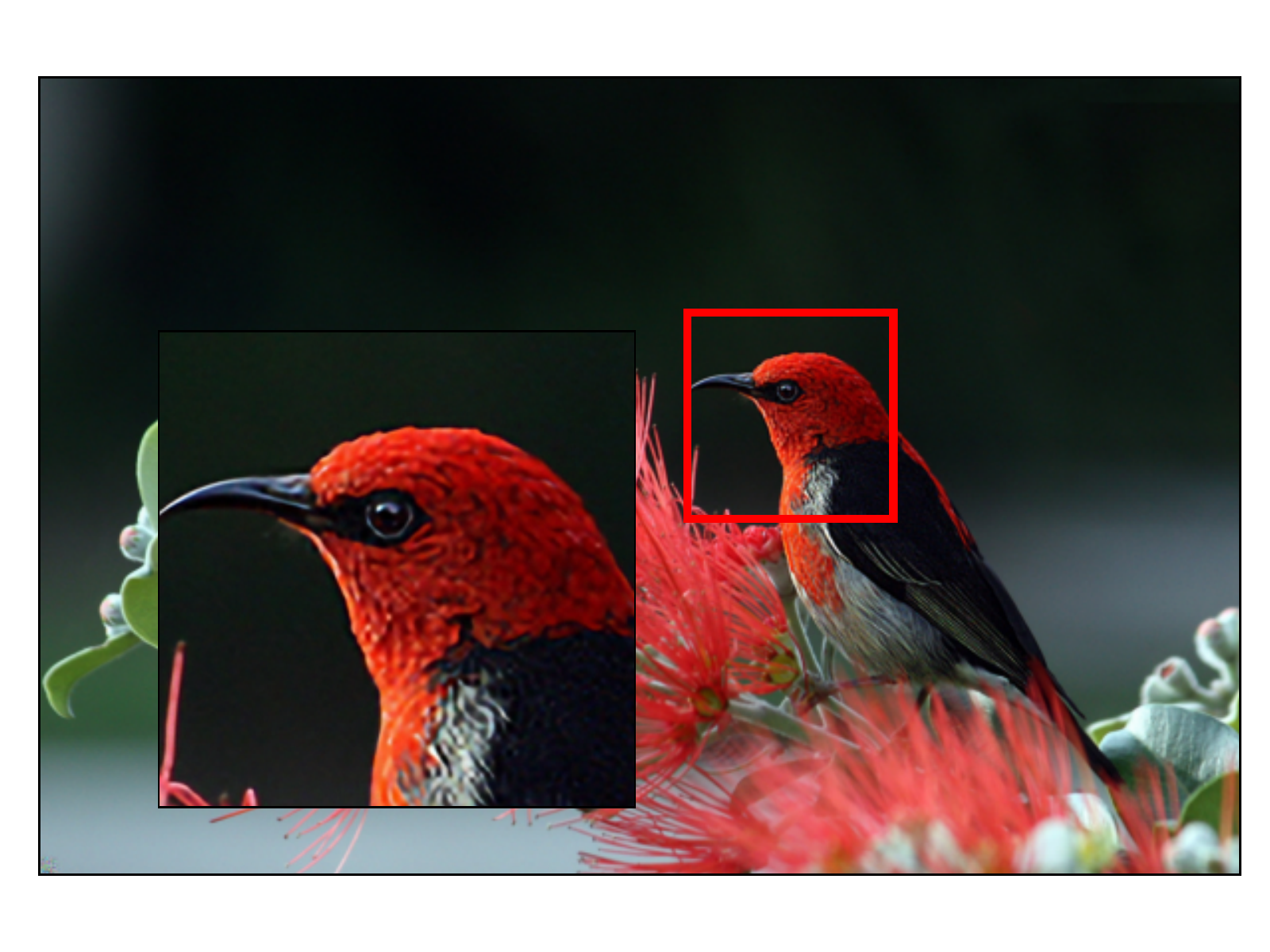}
       \caption{SR}
       \label{fig:subim2}
   \end{subfigure}
\hfill 
   \begin{subfigure}{0.33\textwidth}
       \includegraphics[width=\linewidth]{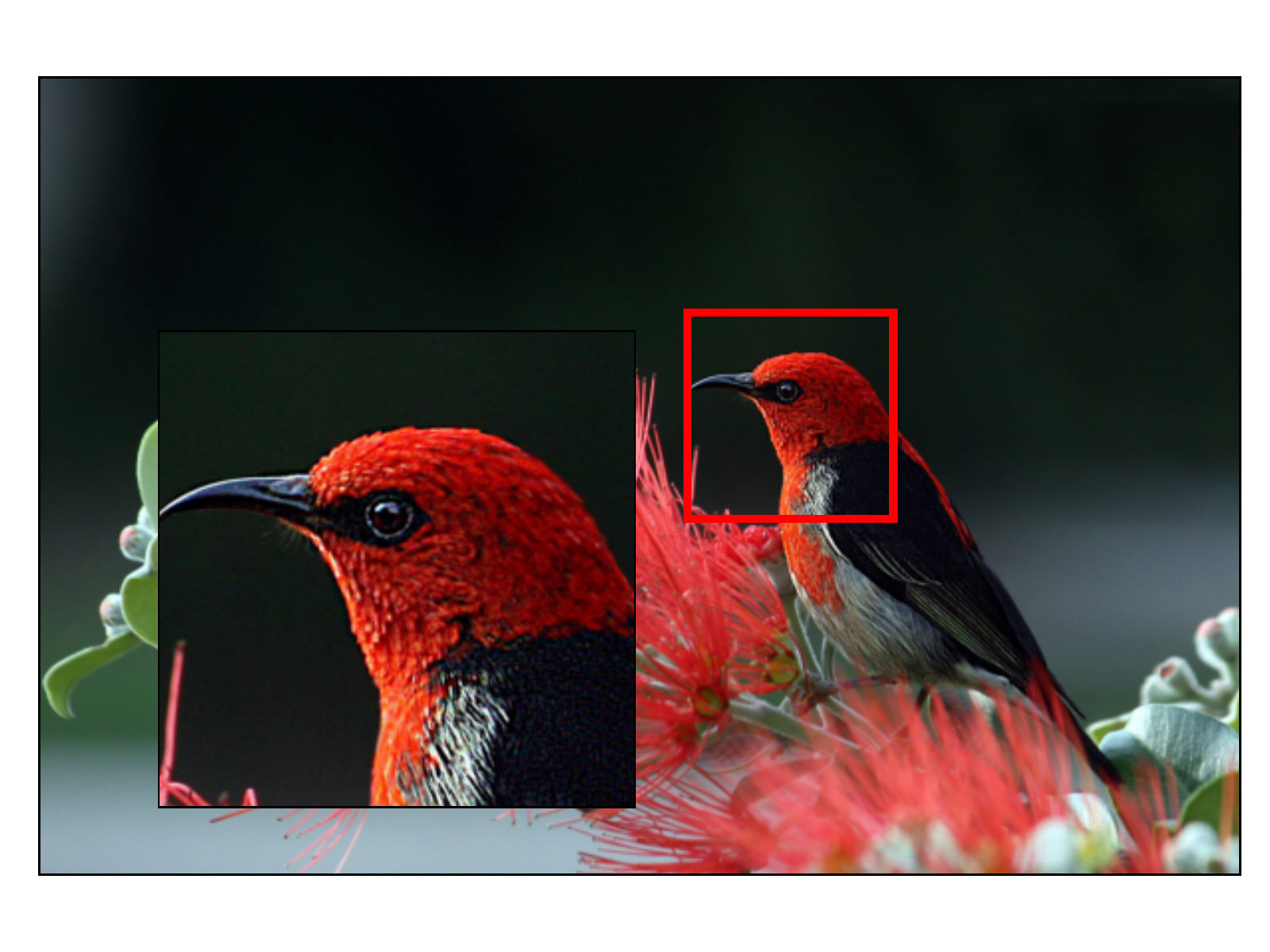}
       \caption{HR}
       \label{fig:subim3}
   \end{subfigure}

   \caption{A side-by-side comparison of a LR, SR, and HR image (4x scaling) from the DIV2K validation set.}
   \label{fig:birddetail}
\end{figure*}

\begin{table*}
  \centering
    \begin{tabular}{l c c c c}
      \toprule
      \textbf{Methods} & \textbf{Set5}& \textbf{Set14} & \textbf{BSDS100}\ & \textbf{General100}\\
      (generative) & \cite{bevilacqua2012low} & \cite{zeyde2010single} & \cite{martin2001database} & \cite{dong2016accelerating}\\
      \midrule
      Bicubic  & 0.3407 &  0.4393 &  0.5249 &  0.3528\\
      SFTGAN \cite{wang2018recovering} & 0.0890 & 0.1481 &  0.1769 & 0.1030\\
      SRGAN \cite{ledig2017photo}  & 0.0882 &  0.1663 &  0.1980 & 0.1055\\
      ESRGAN \cite{wang2018esrgan} & 0.0748 &  0.1329 &  0.1614  & 0.0879\\
      NatSR \cite{soh2019natural} & 0.0939  & 0.1758 & 0.2114 & 0.1117\\
      SPSR \cite{ma2020structure} & \textbf{0.0644}  &  \underline{0.1318}  & \underline{0.1611}  & \underline{0.0863}\\
      DiWa (ours)  & \underline{0.0747} & \textbf{0.1143} & \textbf{0.1398} & \textbf{0.0783}\\
      \bottomrule
      
  \end{tabular}
  \caption{LPIPS Comparison with state-of-the-art generative SR methods for scale x4. }
\label{tab:spsr_results}
\end{table*}

\subsubsection{General Super-Resolution}
\autoref{tab:div2k_results} presents the results of our proposed method for 4x scaling on the DIV2K validation set. 
Unlike in face SR, we did not resize the test images to a fixed size.
We evaluate our method against SRDiff \wrt PSNR, SSIM, and LPIPS. 
Note that \autoref{tab:div2k_results} includes regression-based methods, which result in higher PSNR and SSIM values than generative approaches \cite{saharia2022image}.
After training for 100k steps, our method outperforms all tested state-of-the-art generative methods (RankSRGAN \cite{zhang2019ranksrgan}, ESRGAN \cite{wang2018esrgan}, and SRFlow \cite{lugmayr2020srflow}) and SRDiff in terms of LPIPS and PSNR.
Regarding SSIM, our approach also achieves a competitive score of 0.78, which is 0.01 less than the best result by SRDiff.

A comparison between our approach and state-of-the-art generative approaches on Set5, Set14, BSDS100, and General100 is summarized in~\autoref{tab:spsr_results}. 
Our approach outperforms compared methods \wrt LPIPS on all datasets, except for Set5, where we achieve the second-best result.
Both experiments with low LPIPS show that our approach is more effective at perceptual similarity judgments than other generative approaches.

An example reconstruction image obtained using our method is shown in \autoref{fig:birddetail}.
Together with the quantitative measurements, it shows that our model can refine the input while preserving the texture (see the zoomed-in region).

Regarding convergence speed, SRDiff used 100k iterations to pre-train their encoder structure and trained the denoise function in combination with 300k iterations.
In contrast, we used a small number of 100k iterations without encoder pre-training.
Also, we used 2.7M fewer parameters than SRDiff.
Thus, our model exhibits a faster convergence rate without forfeiting its generalization ability.

Our observation on general SR supports the findings of Wang et al. \cite{wang2020high} regarding the capability of CNNs to exploit high-frequency details for image classification.
They found that a CNN achieving higher classification accuracy exploits more high-frequency components of the input image.
Since LPIPS is calculated with a CNN classifier, effectively hallucinating high-frequency details favors a low LPIPS score for SR.
Hence, our method effectively retains the high-frequency information in the generated HR images, leading to improved performance for CNN classifiers, as evidenced by low LPIPS scores.

We also found that some reconstructions contain small amounts of Gaussian noise in the bottom left corner, which is not apparent in the face SR setting (with fixed spatial size), also present in \autoref{fig:birddetail}. 
We hypothesize that it likely comes from the network architecture's positional encoding.
It might be mitigated by removing the positional encoding or training for a more significant number of iterations.

\begin{table}[!h]
\center
\small
\begin{tabular}{l c c  }
\toprule
\textbf{Methods} & \textbf{PSNR} $\uparrow$ & \textbf{SSIM} $\uparrow$ \\ 
\midrule
SR3 (baseline) & 22.74 & 0.6363 \\
SR3 + 2D-DWT & 26.94 & 0.7150  \\
SR3 + init. predictor & \underline{27.02} & \textbf{0.7540}  \\
SR3 + 2D-DWT + init. predictor & \textbf{27.37} & \underline{0.7220}  \\
\bottomrule
\end{tabular}
\caption{Ablations of our approach for general SR evaluated on Set5 and trained for 100k iterations on DIV2K ($48\times48 \rightarrow 192\times192$). Both components positively impact the baseline, while their fusion combines the strength of both components, resulting in the highest PSNR value.}
\label{tab:abl}
\end{table}

\subsubsection{Ablation Study}
We conduct an ablation study to probe the influences of the initial predictor and the 2D-DWT. 
We evaluated different variations for general SR ($48\times48 \rightarrow 192\times192$), namely Set5, and trained each variation for 100k iterations on DIV2K with a batch size of 64. 
All remaining hyper-parameters are identical to the other experiments.

\autoref{tab:abl} presents the results of our ablation studies. 
They demonstrate that both components,  the initial predictor and the 2D-DWT, positively influence the quality of the final reconstruction individually by a large margin, as measured by PSNR and SSIM. 
Using both components together results in the best PSNR performance, although with a slight decrease in SSIM compared to using only the initial predictor.

It is worth noting that the goal of our method is not to replicate the HR image perfectly but rather to improve the resolution of the SR image as much as possible while preserving high-frequency details. 
Since the SR problem is ill-posed, this goal is shared across current research.
The comparison highlights the trade-offs in generative SR, and our proposed method strikes a balance between preserving fine details and the overall naturalness of the image.

\section{Future Work}
For future work, our method still faces challenges, which are also apparent in SR3, that require further investigation, such as preserving fine skin texture details (e.g., moles, pimples, and piercings).
These limitations, partly due to the ill-posed problem definition, should be addressed when implementing our method in real-world scenarios.
Despite this, it would be interesting to see this approach applied to multi-level DWT, similar to MWCNN \cite{liu2018multi}.
Alongside latent diffusion \cite{rombach2022high, ramesh2021zero, ramesh2022hierarchical, saharia2022photorealistic}, it would broaden the accessibility to experiments like in SR3.
Also, we expect further improvements by using EMA, but this would come with additional training time.
Regarding architectural aspects, further exploration of initial predictors is an exciting direction for future research \cite{liu2018multi, zou2021sdwnet, xue2020wavelet}.

\section{Conclusion}
In this work, we presented a novel Difussion-Wavelet (DiWa) approach for image super-resolution that leverages the benefits of conditional diffusion models and wavelet decomposition. 
We evaluated our approach on two face SR tracks (8x scaling) against SR3.
For general SR (4x scaling), we compared our method against SRDiff and non-diffusion-based, generative approaches.
Our experiments show the effectiveness of our method by outperforming state-of-the-art generative techniques in terms of PSNR, SSIM, and LPIPS for both tasks.

Furthermore, our approach's reliance on the wavelet domain improves inference speed. 
Since the wavelet domain is spatially four times smaller than the image space, our approach benefits from reduced memory consumption and faster processing times. 
Additionally, DWT reduces the required receptive field of the denoise function, which further contributes to its faster inference and convergence speed. 

These optimizations enable our approach to achieve state-of-the-art results while requiring only 92M parameters instead of 550M compared to SR3 and 9.3M instead of 12M compared to SRDiff. 
Therefore, our approach is not only effective but also lightweight, making it an attractive option for researchers with limited access to high-performance hardware. 
It offers high-quality image reconstructions and a practical approach that can be readily reproduced and added to existing diffusion pipelines.

The impact of this work extends beyond the field of SISR. 
With its improved inference and convergence speed, reduced memory consumption, and compact parameter size, our approach is well-suited for widespread adoption in various applications, including real-time SR and pre-processing for downstream tasks like image classification, medical imaging, multi-image SR, satellite imagery, and text-to-image generation \cite{frolov2021adversarial, bashir2021comprehensive, 10041995}.
Also, the low LPIPS evaluations of our method indicate that our method is interesting as a pre-processing step for image classification.

\section*{Acknowledgment}
This work was supported by the BMBF project XAINES (Grant 01IW20005) and by Carl Zeiss Foundation through the Sustainable Embedded AI project (P2021-02-009). 

{\small
\bibliographystyle{ieee_fullname}
\bibliography{main}
}

\end{document}

%% file: algorithms/training.tex
\begin{algorithm}[t]
\caption{DiWa - Training}
\label{alg:training}
\begin{algorithmic}[1]
    \REQUIRE $f_\theta$: Denoiser network, $g_\theta$: Initial predictor, \\$\mathcal{D}$ = $\{\left(\mathbf{x}, \mathbf{y}\right)\}$: LR and corresponding HR image pairs,\\ $\alpha_{1:T}$: Noise schedule.
    \REPEAT
        \STATE $\left(\mathbf{x}, \mathbf{y}\right) \sim \mathcal{D}$ \hfill
        \STATE $\check{\mathbf{x}} \gets \text{2D-DWT} \left( \mathbf{x} \right)$ \hfill
        \STATE $\check{\mathbf{y}} \gets \text{2D-DWT} \left( \mathbf{y} \right)$ \hfill
        \STATE $t \sim \mathrm{Uniform}(\{1,\cdots,T\})$ and $\varepsilon_t \sim \mathcal{N}(\textbf{0},\,\textbf{I})$ \hfill 
        \STATE  $\gamma_t \gets \prod_{i=1}^t \alpha_i$ \hfill
        \STATE $\mathbf{z}_{t} \gets \sqrt{\gamma_t} \left( \check{\mathbf{y}} - g_\theta \left( \check{\mathbf{x}} \right)\right) + \sqrt{1-\gamma_t}\varepsilon_t$
        \STATE Take gradient step on \\ 
        \quad $\nabla_{\theta}\|\varepsilon_t - f_\theta \left( \check{\mathbf{x}}, \mathbf{z}_t, \gamma_t \right)\|$ \\ 
    \UNTIL converged
\end{algorithmic}
\end{algorithm}

%% file: algorithms/inference.tex
\begin{algorithm}[t]
\caption{DiWa - Inference in $T$ refinement steps}
\label{alg:inference}
\begin{algorithmic}[1]
    \REQUIRE $f_\theta$: Denoiser network, $g_\theta$: Initial predictor, \\$\mathbf{x}$: Blurry input image, $\alpha_{1:T}$: Noise schedule.
    \STATE $\check{\mathbf{x}} \gets \text{2D-DWT} \left( \mathbf{x} \right)$
    \STATE $\mathbf{x}_{\text{init}} \gets g_\theta(\check{\mathbf{x}})$ \hfill 
    \STATE $\mathbf{z}_T \sim \mathcal{N}(\textbf{0},\,\textbf{I})$ \hfill 
    \FOR{$t = T, \ldots, 1$}
        \STATE $\varepsilon_t \sim \mathcal{N}(\textbf{0},\,\textbf{I})$
        
        \STATE $\mathbf{z}_{t-1} \gets \mu_{\theta}(\check{\mathbf{x}}, \mathbf{z}_{t}, \gamma_t) + \sqrt{1-\alpha_t}\varepsilon_t$ \\
    \ENDFOR
    \RETURN $\text{2D-IDWT} \left(\mathbf{x}_{\text{init}}  + \mathbf{z}_{0} \right)$
\end{algorithmic}
\end{algorithm}